\documentclass[a4paper, 10 pt, conference]{ieeeconf}  

\IEEEoverridecommandlockouts                              

\usepackage{cite}
\usepackage{amsmath,amssymb,amsfonts}
\usepackage{algorithmic}
\usepackage{graphicx}
\graphicspath{{Images/}}
\usepackage{textcomp}
\usepackage{xcolor}
\usepackage{float}
\usepackage[utf8]{inputenc}
\usepackage[T1]{fontenc}  

\title{\LARGE \bf
A Collaborative Framework for High-Definition Mapping
}

\author{Alexis Stoven-Dubois$^{1}$, Kuntima Kiala Miguel$^{1}$, Aziz Dziri$^{1}$, Bertrand Leroy$^{1}$ and Roland Chapuis$^{2}$
\thanks{$^{1}$Mobility Department, VEDECOM, F-78000 Versailles, France
        {\tt\small alexis.stoven-dubois@vedecom.fr}}%
\thanks{$^{2}$Université Clermont Auvergne, CNRS, SIGMA Clermont, Institut Pascal, F-63000 Clermont-Ferrand, France
        {\tt\small roland.chapuis@uca.fr}}%
}

\begin{document}

\maketitle
\thispagestyle{empty}
\pagestyle{empty}

\begin{abstract}

For connected vehicles to have a substantial effect on road safety, it is required that accurate positions and trajectories can be shared. To this end, all vehicles must be accurately geolocalized in a common frame. This can be achieved by merging GNSS (Global Navigation Satellite System) information and visual observations matched with a map of geo-positioned landmarks. Building such a map remains a challenge, and current solutions are facing strong cost-related limitations.

We present a collaborative framework for high-definition mapping, in which vehicles equipped with standard sensors, such as a GNSS receiver and a mono-visual camera, update a map of geolocalized landmarks. Our system is composed of two processing blocks: the first one is embedded in each vehicle, and aims at geolocalizing the vehicle and the detected feature marks. The second is operated on cloud servers, and uses observations from all the vehicles to compute updates for the map of geo-positioned landmarks. As the map's landmarks are detected and positioned by more and more vehicles, the accuracy of the map increases, eventually converging in probability towards a null error. The landmarks geo-positions are estimated in a stable and scalable way, enabling to provide dynamic map updates in an automatic manner.
\end{abstract}

Collaborative Techniques \& Systems, Accurate Global Positioning, Sensing, Vision and Perception

\section{Introduction}

Intelligent communications will soon be enhancing all standard and autonomous vehicles on the roads. This new development is expected to have a significant impact on road safety, by allowing collective and real-time exchange of positions and observations, including infrastructure status (e.g. construction works) and locations of unplanned events (e.g. road accidents), between vehicles.
Nevertheless, achieving such a safety improvement requires that vehicles and other road data are accurately geolocalized.

Geolocalization performed using GNSS (Global Navigation Satellite System) information cannot guarantee accurate positioning, especially in "urban canyons". Another localization strategy consists in using embedded visual sensors, and matching images over accurately positioned landmarks, hence requiring the preliminary development of a high-definition map containing geolocalized landmarks.
Recently, major actors in the field have tried building high-definition maps by deploying fleets of vehicles equipped with high-end sensors. Having elaborated high-definition maps for the major highways, they are now facing strong logistical and economical limitations, and are not considering to register entire road networks in the near future \cite{Seif2016}. Furthermore, due to the usage of dedicated vehicles fleets equipped with high-end sensors, live update of the maps can not be considered.

Instead of relying on fleets of dedicated vehicles, we intend to efficiently use production vehicles equipped with GNSS receivers and front mono-visual cameras, which will soon be part of the standard equipment. These vehicles, whether they are man- or self-driven, will collaborate in the build-up of a geolocalized map by visually identifying landmarks and measuring their geo-positions. The accuracy of such a map will be the outcome of the crowdsourcing of a vast amount of geo-position measurements. In turn, these vehicles could make use of this map to enhance their safety by accurately geolocalizing themselves within the infrastructure, and sharing accurate positions and trajectories.

The main contribution of this paper is the proposition of a scalable collaborative framework for high-definition mapping, which overcomes the economic lock attached to the making of high-definition maps on large territories, and allows for constant updates of the map. We show that the incremental process of refining landmarks geo-positions through measurements crowdsourcing allows for a refinement of the map accuracy.

This paper is organized as follows: in the next section, we discuss previous methods for mapping areas and geolocalizing the maps. Next, we present the framework for our crowdsourced mapping solution. Then, we show results of the first experiments of our geolocalized map-building application. Finally, we discuss the results and detail next directions for the improvement of our system.

\section{State-of-the-Art}
Connected vehicles and autonomous ones require accurate geolocalization, eventually up to the decimeter-level, for efficient data-sharing \cite{Ziegler2014}. The use of GNSS receivers is not sufficient to achieve such an accuracy, as even more expensive RTK-GNSS technology suffer from multi-path and unavailability issues in urban areas \cite{Bresson2017}. On the other hand, the use of highly accurate landmarks allows the vehicle to geolocalize itself using its embedded sensors, even in urban environments, as shown in \cite{Spangenberg2016} and \cite{Buczko2017}, where a map of pole-shaped landmarks and a map of traffic signs are respectively used.\newline
Those two maps are only built in an unscalable way and designed to be applied within restricted experiments zones. The building of a large map made of accurate geolocalized landmarks remains a real challenge, as the map must be stable to new updates by the different vehicles, and scalable, i.e. have low computations and storage requirements.

Photogrammetry applications seek to build geolocalized maps of various objects or landscapes with the highest possible accuracy \cite{Westoby2012}. To achieve this, feature points are extracted and matched within massive amounts of images, additional sensing outputs are added if available, and a global optimization (bundle adjustment) is operated \cite{Ozyesil2017}. Geolocalized data can be included through the use of GNSS sensors \cite{Rehak2016}, making the accuracy sensitive to GNSS flaws, and by manually installing geolocalized anchor points in the scene \cite{Harwin2012}, which is a tedious and expensive operation. This process must be performed offline, and is unscalable for producing large maps of accurate landmarks.

SLAM (Simultaneous Localization and Mapping) techniques have aimed to build maps as a support for various robots localization and navigation algorithms \cite{Suger2017}. Although the process is similar to the photogrammetry one, filtering \cite{Tiefenbacher2015} or local optimization \cite{Pirker2011}, which provide less accurate maps than with a global optimization, are generally preferred due to real-time requirements. Recently, C-SLAM methods have enabled different vehicles to map and position themselves within a common reference frame. However, such techniques induce intensive communication and computation requirements, and are not suited for building large geolocalized maps. Moreover, SLAM-based maps are often built with regard to the vehicle's reference, and can only be geo-referenced through the use of GNSS sensors, again making the map accuracy sensitive to GNSS weaknesses \cite{Bresson2017}.

Our proposition is an incremental process which crowdsources measurements from standard vehicles to update a map of geolocalized landmarks through a centralized optimization process. Our framework starts with an empty map, and does not require any prior knowledge regarding the landmarks configuration. Collecting observations from a vast amount of vehicles provides dynamic updates of the map, decoupling the localization and mapping warrants stability to communications outages, and strongly selecting the landmarks to register ensures long-term scalability.\newline

\section{System Overview}
\begin{figure*}
\begin{center}
\includegraphics[width=1.0\textwidth]{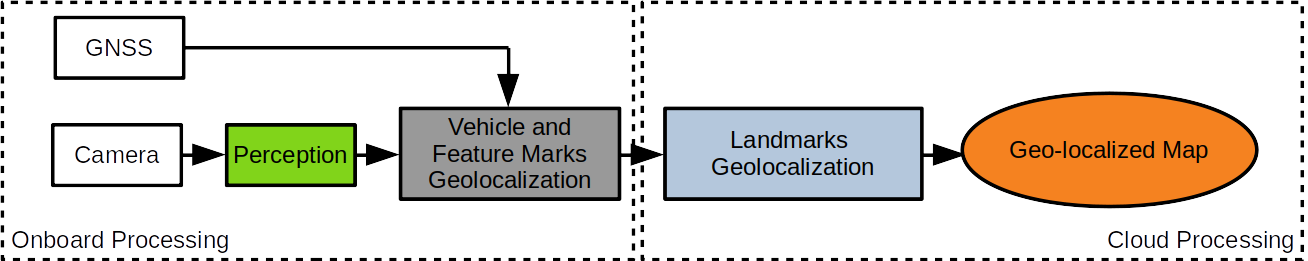}
\caption{Collaborative pipeline - Each vehicle visually detects feature marks. Using its embedded sensors, it geolocalizes both itself and the feature marks. Cloud servers, receiving geo-positioned feature marks from several vehicles, can update the map of geolocalized landmarks.} 
\label{fig:Pipeline}
\end{center}
\end{figure*}

For our framework to be fully cooperative, we consider all connected vehicles equipped with the following set of standard sensors: a GNSS receiver and a mono-visual camera. Fig. \ref{fig:Pipeline} illustrates this framework, which can be split into onboard perception and localization tasks, and a cloud processing.\newline

\subsection{Onboard Processing}
The onboard processing operations described below are implemented on each vehicle to process raw data delivered by the sensors. They aim to localize both the vehicle and the feature marks, and to communicate such information to cloud servers.\newline

\subsubsection{Perception}
The \textit{Perception} block receives images from the camera as inputs, detects and describes feature marks within them, and provides their descriptions as outputs. As these descriptions will be matched on cloud servers with the descriptions of the map's landmarks, they must be robust to variable imaging conditions, including scale and environmental and illumination changes. This ensures that vehicles driving through the same areas detect similar feature marks.

Urban areas include many roadside elements that can be used as feature marks. For this purpose, we have chosen to use traffic signs, as they are semantic objects which can be robustly matched \cite{Botterill2013}, and frequently observed especially in urban environments. For detecting traffic signs, a CNN-derived architecture is used \cite{Zhu2016}, which provides corresponding bounding boxes. Each traffic sign observed in an image is described using both its bounding box pixel-position, and its semantic information. It can be noted that other types of feature marks can be considered simultaneously and added to the \textit{Perception} block in the future.\newline

\subsubsection{Vehicle and Feature Marks Geolocalization}
This block's inputs are the geo-positions from the GNSS receiver and the descriptions of feature marks from the \textit{Perception} block. Based on this information, it computes the geolocalization of both the vehicle and the feature marks, and provides geolocalized feature marks observations as outputs. Those latters are registered on the vehicle and uploaded to cloud servers whenever possible.

\begin{figure}
\begin{center}
\includegraphics[width=0.4\textwidth]{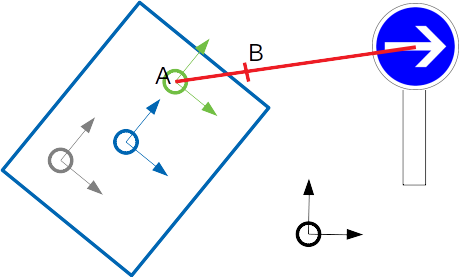}
\caption{2D representation of frames and projection lines (view from the top) - The world frame is depicted (black), along with the vehicle frame (blue), the camera frame (green), and the GNSS receiver frame (grey). The projection line and its two points $A$ and $B$ originating from the detection of the traffic sign are also shown (red).} 
\label{fig:Reperes}
\end{center}
\end{figure}

All vehicles and traffic signs are to be geolocalized within a world frame. The vehicle and its sensors relate to three different frames, as depicted in Fig \ref{fig:Reperes}: the vehicle frame situated at its center, the GNSS receiver frame, and the camera frame. Relations between these frames are obtained from an extrinsic calibration procedure, which provides:
\begin{itemize}
\item The transform matrix $T^{G}_{V}$ linking the vehicle frame to the GNSS receiver frame.
\item The transform matrix $T^{C}_{V}$ linking the vehicle frame to the camera frame.
\end{itemize}

The true states $X^{V}$, $X^{G}$ and $X^{C}$ of the vehicle, GNSS receiver and camera are defined by their respective geo-positions and heading.
The GNSS receiver provides an observation $Z^{G}$ of its state $X^{G}$, including both its geo-position and heading (as the vehicle is moving):
\begin{equation}
Z^{G} = g(X^{G}, w)
\end{equation}
with $g$ being the GNSS receiver observation model, and $w$ being its noise.
Making use of $T_{V}^{G}$ linking the GNSS frame to the vehicle frame, the estimation $\hat X^{V}$ of the vehicle state is computed directly from the GNSS observation $Z^{G}$.
As for the camera, an estimation $\hat X^{C}$ of its state is computed from $\hat X^{V}$, making use of $T_{V}^{C}$ linking the camera frame to the vehicle frame.

Each traffic sign description is associated with an observation of its bounding box pixel-position $Z^{D} = \begin{pmatrix} u & v\\\end{pmatrix}^{T}$. Knowing the camera intrinsic calibration matrix $K$, a projection line linking the camera center to the traffic sign, as depicted in Fig. \ref{fig:Reperes}, can be established and modeled as passing through two points $A$ and $B$ \cite{Ozyesil2017}:
\begin{equation}
A = \hat X^{C}
\end{equation}
\begin{equation}
B = \hat X^{C} + (T_{W}^{C})^{-1}  K^{-1}  \begin{pmatrix}
u & v & 1\\
\end{pmatrix}^{T}
\end{equation}
with $T_{W}^{C}$ being the transform matrix, computed from $\hat X^{C}$, which links the camera frame to the world frame. The state $X^{S}$ of a traffic sign is defined by its geo-position, thus each projection line $(A, B)$ originating from a detection of a sign constitutes an observation $Z^{S}$ of its geo-position.

Finally, the traffic sign observation $Z^{S}$, along with the traffic sign description, is uploaded to cloud servers.

\subsection{Cloud Processing}
The cloud servers receive feature marks observations $Z^{S}$ as inputs, use them to update and improve the map of geolocalized landmarks, and provide map updates as outputs. Such map-building operation does not have to be processed in real-time, but has to be stable to new updates and scalable. No prior knowledge on the map is required, as map updates are able to initialize the map on their own.

First, feature marks observations are matched with the map's landmarks, by comparing their respective descriptions. Our decision to focus on the detection of traffic signs greatly facilitates this task, as the descriptions contain strongly distinguishable semantic information \cite{Botterill2013}. Furthermore, the limited number of traffic signs warrants that this operation remains scalable.

\begin{figure}
\begin{center}
\includegraphics[width=0.5\textwidth]{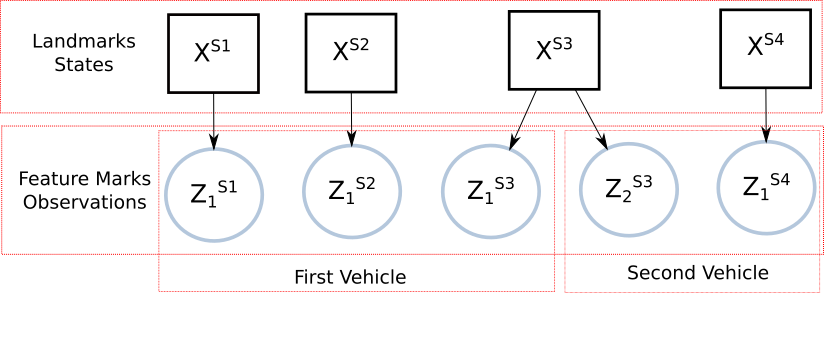}
\caption{Landmarks geolocalization - The cloud servers have received feature marks observations from two different vehicles, and have matched them with various landmarks in the map. The known observations are represented as round factors, the states to estimate as square factors, and the joint constraints as directed arrows.}
\label{fig:Mapping}
\end{center}
\end{figure}

Then, the map is updated, using the feature marks observations to either confirm or infirm the landmarks geo-positions in the map. This operation can be modeled as an optimization problem, as shown in Fig. \ref{fig:Mapping}, with:
\begin{itemize}
\item The traffic signs geo-positions $X^{S_{i}}$ to estimate for each traffic sign $i$.
\item The observations $Z^{S_{i}}_{j}$ of the traffic signs geo-positions, with $Z^{S_{i}}_{j}$ being the $j^{th}$ projection line associated with the traffic sign $i$.
\end{itemize}
Considering a traffic sign $i$, all of its previous and new observations $Z^{S_{i}}_{j}$ are registered and used within the optimization. The new best estimation $\hat X^{S_{i}}$ for the traffic sign geo-position can be obtained through a least-squares triangulation-based optimization, minimizing:
\begin{equation}
\hat X^{S_{i}} = \operatorname*{argmin}_{X^{S_{i}}} \sum_{j} dist(X^{S_{i}}, Z^{S_{i}}_{j})
\label{Regular_Optimization}
\end{equation}
with $dist(X^{S}, Z^{S})$ being the orthogonal distance between the geo-position $X^{S}$ and the projection line $Z^{S}$.

\section{Experiments}
\label{sec:First experiments}
To assess the effectiveness of the proposed solution, a two-step experimentation was performed. First, we validate our solution through a simulation. Then, we confront it to real data.

\subsection{Simulation}
Each vehicle detecting a traffic sign $i$ provides observations $Z^{S_{i}}_{j}$ of its geo-position. Both the detection of the traffic sign and the vehicle geolocalization suffer from some noise, leading to noisy projection lines $Z^{S_{i}}_{j}$. However, as more and more measurements are received, cloud servers are able to estimate the geo-position $\hat X^{S_{i}}$ of the traffic sign with an increasing accuracy, eventually converging towards a null error. To validate this, a simulation of our solution was implemented, without any loss of generality, as a 2D simulation on the North-East plane.

\begin{itemize}
\item First, a traffic sign $i=1$ and its true position $X^{S_{1}}$ are defined along a straight road.
\item For a chosen number $n$ of different vehicle passings, vehicles true states $X^{V}$ are generated.
\item For each vehicle state $X^{V}$, an image is associated, and the corresponding true pixel-position $X^{D}$ for the bounding box of the traffic sign is computed.
\item Next, GNSS observations $Z^{V}$ are generated by applying a random, white noise of $5.0\:m$ for the positions, and $0.35\:rd$ for the orientations, around the vehicles states $X^{V}$.
\item Similarly, feature marks descriptions $Z^{D}$ outputted from the \textit{Perception} block are generated applying a random, white noise of $5\:pixels$ around the pixel-positions $X^{D}$.
\item GNSS observations $Z^{V}$ and feature marks descriptions $Z^{D}$ are fed as inputs for the \textit{Vehicle and Feature Marks Geolocalization} block, which establishes feature marks observations, i.e. projection lines $Z^{S_{1}}_{j}$.
\end{itemize}

In real conditions, the error associated to GNSS measurements is not white and can be affected by strong biases, due to atmospheric conditions and multi-path issues \cite{Tao2015}. Nevertheless, our collaborative approach estimates the landmarks geo-positions using many different measurements acquired during a large time span (with different atmospheric conditions), and by different vehicles with slightly different positions on the road (leading to different multi-path effects). Therefore, we can consider that our assumption for a white distribution of the error still stands.

At each vehicle passing, the \textit{Landmarks Geolocalization} block is activated, and an estimation $\hat X^{S_{1}}$ of the traffic sign geo-position is computed. During the optimization, all feature marks observations $Z^{S_{1}}_{j}$ received at the current vehicle passing and at the previous ones are used. Within our simulation, we simplified the function to optimize:
\begin{equation}
\hat X^{S_{1}} = \operatorname*{argmin}_{X^{S_{1}}} \sum_{j} dist(X^{S_{1}}, Z^{S_{1}}_{j})
\end{equation}
with:
\begin{equation}
dist(X^{S}, Z^{S}) = || head(proj(X^{S}, Z^{S})) - head(Z^{S}) ||
\end{equation}
where $head(Z^{S})$ is the heading (angle to the North axis) of $Z^{S}$, and $proj(X^{S}, Z^{S})$ is the projection line passing through $X^{S}$ and the camera position $\hat X^{C}$ associated with $Z^{S}$. Such a simplification enabled us to estimate at each vehicle passing, not only the traffic sign geo-position $\hat X^{S_{1}}$, but also its covariance $\Sigma^{S_{1}}$ and deviations $\sigma^{S_{1}}$, by applying directly the method from \cite{Eudes2009}.

\begin{figure*}
\begin{center}
\includegraphics[width=1.0\textwidth]{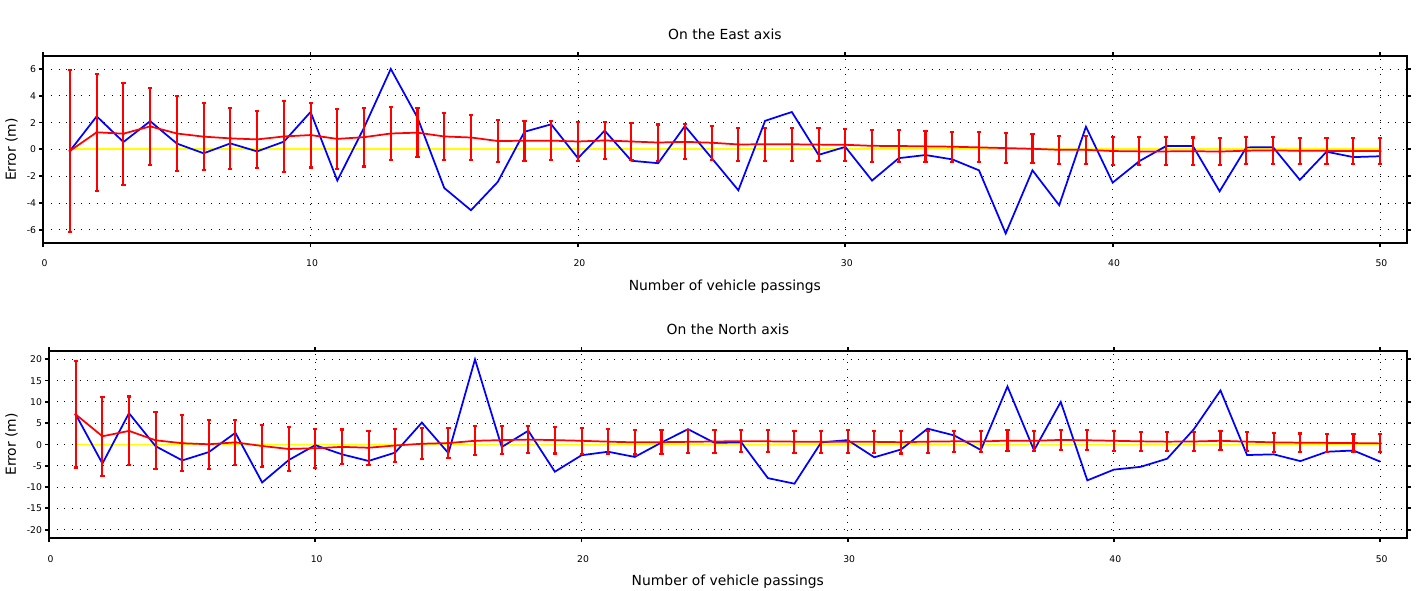}
\caption{Simulation results - Errors on the $East$ and $North$ axes for single-passing measurements (blue) and for estimations of our collaborative approach (red) are shown, as well as the groundtruth (i.e. $error=0$, yellow). Deviations related to the estimations of our solution (red) are also depicted as  $[-2\sigma^{S_{1}};2\sigma^{S_{1}}]$ ranges.}
\label{fig:Simulation_Results}
\end{center}
\end{figure*}

Results of our simulation are shown in Fig. \ref{fig:Simulation_Results}. For each vehicle passing, a single-passing measurement is also computed using only projection lines $Z_{j}^{S_{1}}$ from that passing in the optimization described upper. Errors on the East and North axes for single-passing measurements and for estimations of our collaborative approach are computed as simple differences from the groundtruth $X^{S_{1}}$ , and shown. Also, deviations related to the estimations of our approach are depicted as $[-2\sigma^{S_{1}} ; 2\sigma^{S_{i}}]$ ranges centered around our solution’s estimations.

The results of this simulation confirmed the theoretical assumption of convergence as we see that the deviations on the $East$ and $North$ axes effectively decrease as the number of vehicle passings increases, and that they even decrease with the square root of the number of observations. This indicates that, as more observations are received, the estimation of the traffic sign geo-position converges towards a specific location. Furthermore, we observe that the groundtruth (i.e. $error = 0$) is always comprised within the deviations ranges, indicating that the estimation of the traffic sign geo-position effectively converges towards its true location, hence towards a null error.

\subsection{Real Experiments}
Having verified the convergence hypothesis of our crowdsourced approach through a simulation, we also performed a field-experiment to confirm the effectiveness of the solution in real conditions. The experiment consisted in driving a vehicle, equipped with a standard GNSS sensor and a front-looking mono-visual camera, on a 4 km loop within 4 hours, allowing to acquire data for 10 vehicle passings on the loop. Data from the 10 vehicle passings was shuffled randomly, so as to avoid any bias due to the drawing order. Furthermore, the positions of 10 traffic signs along the loop were acquired manually using an RTK-GPS receiver, providing a groundtruth with which the results of our approach have been compared.

For each considered traffic sign $i=1,...,10$, a number of detections occured at each vehicle passing, leading to feature marks descriptions outputted from the \textit{Perception} block. Feeding both those and GNSS observations $Z^{V}$ as inputs for the \textit{Vehicle and Feature Marks Geolocalization} block, feature marks observations (projection lines) $Z^{S_{i}}_{j}$ are outputted. Finally, at the end of each vehicle passing, the regular optimization of the \textit{Landmarks Geolocalization} block is processed, giving an estimation $\hat X^{S_{i}}$ of the geo-position of each traffic sign $i$.

\begin{figure*}
\begin{center}
\includegraphics[width=1.0\textwidth]{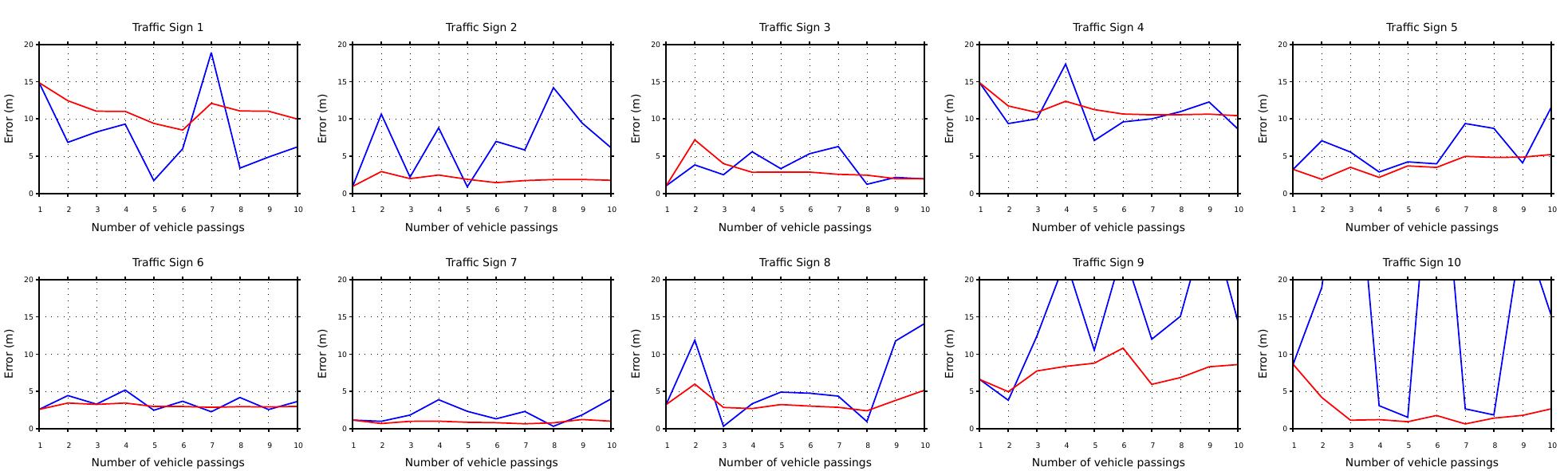}
\caption{Field-tests results: Errors for single-passing measurements (blue) and for estimations of our collaborative approach (red). Errors for single-passing measurements of the traffic signs $i=9$ and $i=10$ are high in some cases. For visibility purposes, they are not depicted.
}
\label{fig:Real_Results_All_Pannels}
\end{center}
\end{figure*}

The results obtained for the different traffic signs are depicted in Fig. \ref{fig:Real_Results_All_Pannels}. As previously for the simulation, single-passing measurements are also computed at each vehicle passing using only the feature marks observations $Z^{S_{i}}_{j}$ from that passing in the regular optimization. Errors for estimations of our collaborative approach and for single-passing measurements are computed as distances from the groundtruth $X^{S_{i}}$, and shown.

The results show that single-passing measurements for the traffic signs geo-positions may be extremely inaccurate in some cases. This corresponds to a low quality of vehicle geolocalization provided by the GNSS receiver, as our vehicle was driven in a condensed urban zone. In the meanwhile, estimations of our collaborative approach quickly surpass the average accuracy of single-passing measurements, indicating that our crowdsourced approach gives better performances. However, the convergence in probability of the error confirmed by simulation is not observable here, due to the small amount of vehicle passings acquired during the tests.

\begin{figure}
\begin{center}
\includegraphics[width=0.5\textwidth]{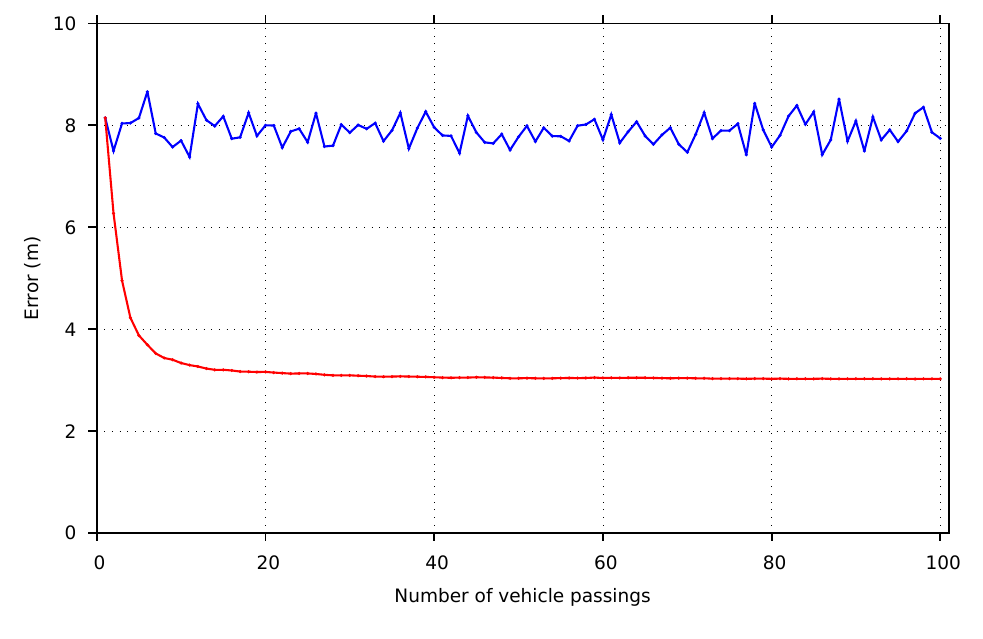}
\caption{Average errors for single-passing measurements (blue) and for estimations of our collaborative approach (red), when observations of all the traffic signs are superposed on the first traffic sign, generating 100 vehicle passings.}
\label{fig:All_Pannels}
\end{center}
\end{figure}

In order to observe the convergence using the data acquired during the field-test, we propose to build a dataset according to the two following principles:
\begin{itemize}
\item We take the first traffic sign as a reference, and translate all the geo-positions observations $Z^{S_{i}}_{j}$ of the other traffic signs $i=2,...,10$ to the reference geo-position. This allows to transform the 10 traffic signs with $10$ passings into $1$ traffic sign observed during $100$ passings.
\begin{equation}
t_{l}^{1} = t(X^{S_{l}} - X^{S_{1}})\:\:;\:\: l=2,...,10
\end{equation}
where $t_{l}^{1}$ is the translation allowing to superpose observations of the traffic sign $i=l$ to the reference traffic sign $i=1$. $X^{S_{l}}$ is the groundtruth position of the traffic sign $i=l$.
\item Furthermore, to avoid any bias due to drawing order, we compute 1000 permutations of these 100 vehicle passings, and average at each passing the error resulting of the geo-position estimation of our collaborative approach.
\end{itemize}
The results obtained using this new dataset are depicted in Fig. \ref{fig:All_Pannels}, and show that our collaborative approach outperforms single-passing measurements. As the number of vehicle passings increases, our method converges to an error of 3 meters while single-passing measurements provide an average error of 8 meters. The magnitude of this error can be explained by an imprecision of the setup of the camera and GNSS receiver on the test-vehicle, due mostly to an inaccuracy of synchronization and extrinsic calibration of the camera with regard to the GNSS receiver. Nevertheless, this experiment confirms the hypothesis of convergence in probability and shows the effectiveness of the proposed approach.
 
\section{Conclusion}
In this paper, we have presented a collaborative mapping framework which will allow all connected vehicles equipped with a GNSS receiver and a mono-visual camera to position landmarks with accuracy. The map is collaboratively built out of landmark geo-position measurements performed by standard vehicles. This crowdsourcing approach, while being scalable and cost-effective, allows for continuous updates of the map. Finally, a field-test has been presented, showing that our collaborative approach has a better positioning accuracy in average than independent geo-position measurements.
While field experiments are still going on in order to gather more data, several enhancements can be envisioned:
\begin{itemize}
\item Our proposition only considers traffic signs for landmark detection. This could be generalized to other types of features marks and objects, which will be especially useful in areas where traffic signs are scarce.
\item The present implementation assumes that the map's landmarks are manually selected during initialization. An enhancement could consist in implementing a dynamic management of the feature marks, where new landmarks could be proposed by vehicles and obsolete ones could be revoked by the cloud servers. Doing so will allow to automatically obtain a higher density of landmarks, and result in a more efficient updating of the map.
\end{itemize}

\addtolength{\textheight}{-12cm}   


\bibliographystyle{ieeetr}
\bibliography{references}

\end{document}